%% file: main.tex
\documentclass[10pt,twocolumn,letterpaper]{article}

\usepackage[pagenumbers]{cvpr} 

\input{preamble}

\definecolor{cvprblue}{rgb}{0.21,0.49,0.74}
\usepackage[pagebackref,breaklinks,colorlinks,citecolor=cvprblue]{hyperref}

\title{Contrastive Learning-Based Spectral Knowledge Distillation for Multi-Modality and Missing Modality Scenarios in Semantic Segmentation}

\author{Aniruddh Sikdar\footnotemark[1] $\ ^{1}$ , Jayant Teotia\footnotemark[1] $\ ^{1}$, Suresh Sundaram$^{2}$        
\\  
$^{1}$Robert Bosch Centre for Cyber-Physical Systems, Indian Institute of Science, Bengaluru, India\\
$^{2}$Department of Aerospace Engineering, Indian Institute of Science, Bengaluru, India\\
{\tt\small \{aniruddhss, jayantteotia, vssuresh\} @iisc.ac.in}   
}

\begin{document}
\maketitle

\renewcommand{\thefootnote}{\fnsymbol{footnote}}
\footnotetext[1]{Equal contribution.}
\input{sec/0_abstract}    
\input{sec/1_intro}

{
    \small
    \bibliographystyle{ieeenat_fullname}
    \bibliography{main}
}


\end{document}

%% file: preamble.tex
\usepackage[dvipsnames]{xcolor}


%% file: sec/0_abstract.tex
\begin{abstract}
Improving the performance of semantic segmentation models using multispectral information is crucial, especially for environments with low-light and adverse conditions. Multi-modal fusion techniques pursue either the learning of cross-modality features to generate a fused image or engage in knowledge distillation but address multi-modal and missing modality scenarios as distinct issues, which is not an optimal approach for multi-sensor models. To address this, a novel multi-modal fusion approach called CSK-Net is proposed, which uses a contrastive learning-based spectral knowledge distillation technique along with an automatic mixed feature exchange mechanism for semantic segmentation in optical (EO) and infrared (IR) images. The distillation scheme extracts detailed textures from the optical images and distills them into the optical branch of CSK-Net. The model encoder consists of shared convolution weights with separate batch norm (BN) layers for both modalities, to capture the multi-spectral information from different modalities of the same objects.  A Novel Gated Spectral Unit (GSU) and mixed feature exchange strategy are proposed to increase the correlation of modality-shared information and decrease the modality-specific information during the distillation process. 
Comprehensive experiments show that 
CSK-Net surpasses state-of-the-art models in multi-modal tasks and for missing modalities when exclusively utilizing IR data for inference across three public benchmarking datasets. 
For missing modality scenarios, the performance increase is achieved without additional computational costs compared to the baseline segmentation models.

\end{abstract}

%% file: sec/1_intro.tex
\section{Introduction}
\label{sec:intro}


\begin{figure*}[ht]
    \centering
    \includegraphics[scale=0.32]{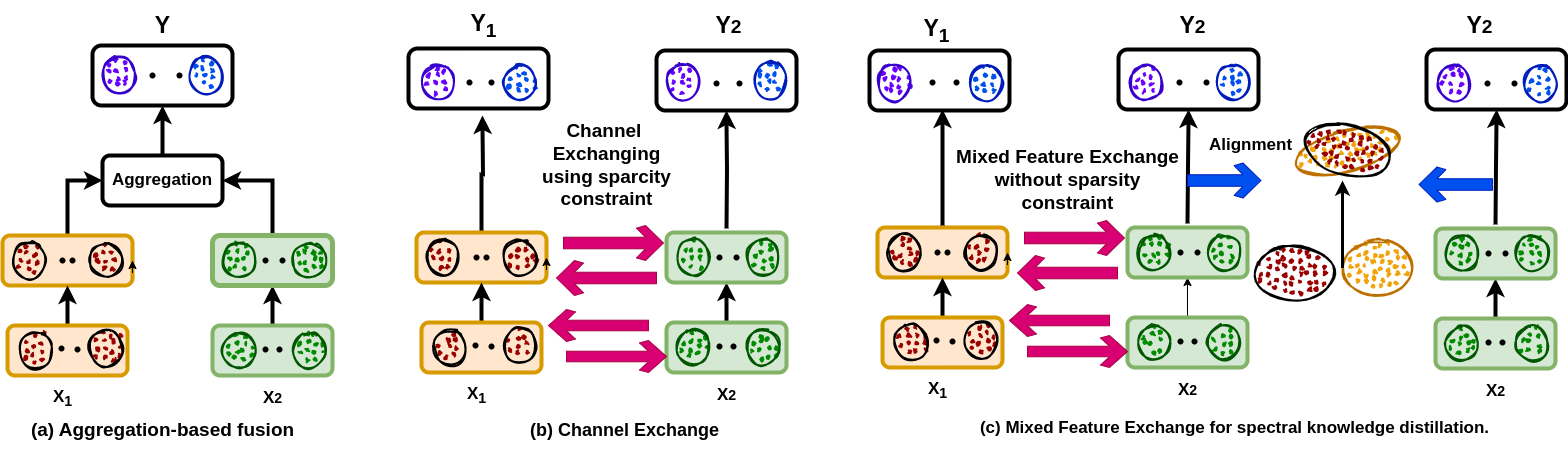}
    \caption{Proposed knowledge distillation framework encourages CSK-Net to transfer semantic knowledge between optical and IR modalities.}
    \label{fig:aggregation}
\end{figure*}

Remarkable advancements have been achieved in scene parsing, leading to high segmentation performance \cite{yuan2020object,zhao2021automatic,zhou2022edge}, however, these advancements are primarily tailored for optical (EO) images. Nonetheless, there is still a critical need to improve the models' generalization capabilities, particularly in adverse conditions like rain, low light, or overexposure. In environments with limited visibility and poor lighting conditions, the adoption of infrared wave (IR) cameras has gained prominence since infrared waves (IR) carry distinct spectral information and have the capability to penetrate dust and smoke\cite{gade2014thermal}. Given the increased accessibility of inexpensive IR sensors, deep multimodal fusion emphasizes the integration of optical and IR modalities \cite{ha2017mfnet,kutuk2022semantic,zhang2021abmdrnet} to leverage advantages over unimodal approaches for semantic segmentation \cite{valada2020self}. 

Multi-modal fusion techniques for EO and IR modalities can be broadly classified into two sub-categories: \textit{Multi-Modality Image Fusion} (MMIF) and \textit{Multi-Modal Feature-level Fusion} (MMFF) techniques.  MMIF techniques aim to generate fused images by modeling the cross-modality features from different sensors.  A common pipeline using Auto-Encoders to fuse optical and infrared images has shortcomings like the shared encoders neglecting modality-specific details\cite{sener2018multi}, missing out on global context to generate high-quality fused images, and losing out the high-frequency information \cite{liang2021swinir,ma2019fusiongan,zhao2021efficient}. The low frequencies represent the information common to both modalities, whereas high frequencies represent the distinctive characteristics specific to each modality, i.e., visible images containing texture details and infrared images displaying thermal radiation. CDDFuse aims to leverage the correlation between low and high frequencies within the image space to constrain the solution space. Despite the superior quality metrics displayed in the fused images of MMIF models, the limitations imposed on these correlations are not fully examined in the training process of segmentation models, leading to performance degradation. \\
To deal with the MMIF model's challenges in handling missing modality scenarios, MMFF techniques involving feature-level fusion, and knowledge distillation are explored to align different sensor modality distributions interactively for inter-modal fusion within the feature space.  For feature-level fusion, studies illustrate the importance of interaction between homogeneous and heterogeneous features \cite{wang2022channel,zhang2023lightweight}, to learn mutual information from dual branches for both modalities. Effective aggregation-based solutions have been proposed using cross \cite{yu2020deformable,wu2021feature} and self-attention \cite{chen2022mixformer} as shown in Fig. \ref{fig:aggregation} (a), but it's been observed that the learnable exchange method does not exhibit superior performance compared to the parameter-free unlearnable exchange \cite{wang2022channel}. When dealing with different sensors, convolutional layers retain the modality-shared features, while the modality-specific information is encoded in the batch norm layers\cite{zheng2021deep}. CEN \cite{wang2022channel} facilitates message passing in the dual branches by dynamically exchanging channels to enable the integration of information, as depicted in Fig. \ref{fig:aggregation} (b). The concept behind CEN is based on the assumption of smaller-norm-less-informative, commonly used in-network pruning \cite{ye2018rethinking}. The sparsity constraint aimed at identifying redundant channels imposes strict constraints that often result in sub-optimal solutions.\\
Pixel-level knowledge distillation strategy \cite{hinton2015distilling} has gained significant attention for training compact models in the context of semantic segmentation tasks, and is used to distill semantic knowledge from optical modality to other modalities. DisOptNet employs a distillation strategy aimed at training the modality-specific branches while simultaneously guiding the optical branch of the model to imitate the feature generation process of a pre-trained optical network for missing modality scenarios. However, directly aligning these images and the distributions of modality-specific features with substantial domain gaps can lead to negative transfer, primarily due to the enforced feature alignment. Most previous works investigate the challenges posed by multi-modal data and missing modality issues as distinct and separate problems.

To address this, our paper proposes a novel \textbf{C}ontrastive learning-based \textbf{S}pectral \textbf{K}nowledge distillation using a mixed feature exchange known as CSK-Net. It consists of shared encoders with distinct batch norms that project the rich semantic and spectral knowledge pertaining to the same object across both modalities into a unified latent space, with the objective of acquiring the modality invariant and modality-specific feature representation from cross-modal data considering diverse perspectives. A mixed feature exchange strategy is proposed using channel and spatial exchange without any sparsity constraints which is simple and self-adaptive, introduces minor randomness in the feature space, and mitigates the forced feature alignment during distillation, as shown in Fig \ref{fig:aggregation} (c). Traditional multimodal analysis systems input two or more sources of diverse data modalities, like video, audio, and images, all of which collectively describe a specific object. Based on this intuition, a novel module called Gated Spectral Fusion is proposed to combine the spectral information from multiple imaging modalities for efficient knowledge distillation. 
Finally, during the early stages of the model, distillation in the shared encoders increases the correlation among low-frequency features. However, to control the correlation of high-frequency features, distillation solely occurs within the modality-specific layer of the optical branch of CSK-Net. This process is supplemented by the use of GSU.
\begin{itemize}
  \item An end-to-end model CSK-Net is proposed to facilitate multispectral semantic 
  segmentation for both multi-modal fusion and missing modality scenarios.   Spectral knowledge distillation is used to distill multi-level semantic features of RGB images into the optical branch of the model, while contrastive learning is used to ensure intra-class compactness and preservation of modality-specific style information.
  \item A novel Gated Spectral Unit and Mixed Feature Exchange strategy is used, to regulate the constraints imposed on the correlation of both low and high-frequency information throughout the distillation process.
  \item A feature reuse strategy is adopted to avoid additional computational costs for missing modality scenarios. This results in the same computation complexity as the baseline segmentation model, with increased performance for infrared images.
  \item Experimental evaluations are conducted for semantic segmentation tasks on three public benchmarking datasets. It demonstrates that CSK-Net consistently shows superior performance to state-of-the-art multimodal fusion and missing modality methods, especially in challenging conditions such as low light and adverse weather conditions.
\end{itemize}

\section{Related Work}

\subsection{Knowledge Distillation}

Knowledge distillation \cite{hinton2015distilling} was initially introduced primarily to transfer knowledge from a complex neural network to a smaller one by minimizing the discrepancy in classification performance between the two models. A knowledge distillation framework named double similarity distillation (DSD) was proposed to increase the classification accuracy, by capturing similarity knowledge in both pixel and category dimensions.
Furthermore, a pixel-wise Similarity Distillation (PSD) module, was designed to capture more intricate spatial dependencies within the data\cite{feng2021double}. Distillation techniques have been explored as a solution to cope with the challenge posed by missing modalities \cite{garcia2018modality,crasto2019mars}. DisOptNet focuses on distilling comprehensive semantic information from the optical modality to SAR, mainly for missing modality scenarios.  MMA-Net introduces a  framework tailored for multi-modal learning. Within this framework, the teacher network is entrusted with transferring comprehensive multimodal information to the deployment network. Concurrently, the regularization network focuses on guiding the deployment network to maintain a balanced approach when dealing with weak modality combinations. This mechanism drives the deployment network to adaptively enhance its capacity in representing the weaker modality combinations.

\subsection{Contrastive Learning}
Contrastive learning has found extensive application in learning representations when labeled data is unavailable \cite{noroozi1603unsupervised}. It is designed to assist in the learning process of distinct feature representations by discerning between similar feature pairs and dissimilar (negative) pairs. In the positive pair sampling strategy, robust perturbations are applied to produce varied perspectives \cite{chen2020simple}. Conversely, negative pairs can be created through random selection or more sophisticated methods such as negative mining \cite{chen2020simple}. PiPa \cite{chen2023pipa} focuses on augmenting intra-image pixel-wise correlations and ensuring patch-wise semantic consistency across diverse contexts. By doing so, it aims to foster intra-class compactness and enhance inter-class separability. 
In one of the seminal works employing contrastive learning for knowledge distillation\cite{tian2019contrastive}, a contrastive-based objective was used with the objective function encouraging both the teacher and student models to map identical inputs to comparable representations.

\subsection{Multi-modal Fusion}


Multi-modal fusion techniques combine information from diverse modalities to capture and consolidate cross-modality features.
Cross-modality transformer \cite{qingyun2021cross} is designed to obtain extensive dependencies across data and incorporate global contextual information throughout the feature extraction process.
In CDDFuse, Restormer blocks are employed to extract low-level features from both modalities. Moreover, a dual-branch transformer is implemented to facilitate long-range attention, effectively managing global features. Furthermore, the integration of Invertible Neural Network (INN) blocks is utilized to extract high-frequency local information within the model. Channel-Exchanging-Network (CEN) \cite{wang2022channel} proposed a dynamic swapping of channels between sub-networks as a mechanism for fusing information from various modalities. This process is self-directed and involves evaluating the significance of individual channels by assessing the magnitude of the Batch-Normalization (BN) scaling factor during the training phase. Multi-spectral segmentation models employ basic fusion strategies, resulting in a reduced discriminability of the fused features. In response to this challenge, a strategy known as "bridging-then-fusing" was introduced \cite{zhang2021abmdrnet}. This approach leverages a bi-directional image-to-image translation method to bridge the gaps between various modalities present in multi-modal data. It subsequently adapts by selectively choosing discriminative multi-modal features.
SegMiF comprises a cascade structure composed of a fusion network and a segmentation network. By linking intermediate features, knowledge can be obtained from the segmentation task effectively aiding the fusion task. A hierarchical interactive attention block is established to ensure precise mapping of crucial information between the two tasks at a fine-grained level to achieve this.
Recently, the denoising diffusion probabilistic model (DDPM) has been applied to the fusion task, functioning as a conditional generation problem embedded within the DDPM framework.

\begin{figure*}[ht]
    \centering
    \includegraphics[scale=0.46, height=4.5cm]{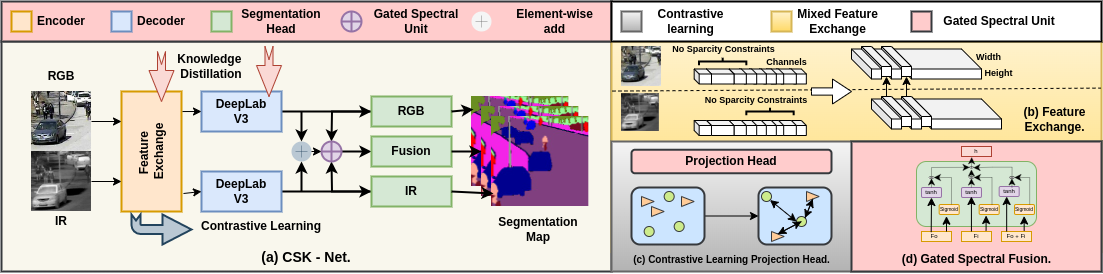}
    \caption{Spectral-based knowledge distillation network (CSK-Net) architecture. (a) The encoders have shared convolutional weights, with individual batchnorms. The output of the decoder block of the two branches is fed into Gated Spectral Unit (GSU) for feature fusion. Pixel-wise contrastive loss is used for features from the four encoder blocks (two from shared and two from IR branch).(b) Encoders are divided into sub-parts, where feature exchange is performed. (c) Pixel-wise contrastive learning loss. (d) Gated Spectral Unit takes three inputs from the Optical branch, IR branch and pixel-wise addition of the features of the two branches and outputs a single feature map.}
    \label{fig:MRFP}
\end{figure*}


\subsection{Problem Formulation}
Thermal imagery holds significance in low-light scenarios, however, deep-learning models experience a drop in performance when exclusively trained on infrared (IR) images, mainly because they contain less semantic information than optical images. The aim is to enhance the model's representation ability across various modalities during training and to maintain this knowledge when performing inference on multi-modal and missing modality settings. Let \{$X^{O},X^{I}$\} = \{($X_{1}^{O},X_{1}^{I}$), ..., ($X_{n}^{O},X_{n}^{I}$)\} denote the co-registered EO-IR image pairs from dataset \textit{D}, with their corresponding pixel-wise labels Y = \{$Y_{1}, .. Y_{n}$\}. Data from both modalities, i.e., \{$X^{O}, X^{I}$\}, are passed as input to the semantic segmentation model $\textit{f}_\theta$ during the training process, where $\theta$ represents the learnable parameters. During inference, for multi-modal settings, \{$X^{O}, X^{I}$\} images are passed to the model, and for missing modality scenarios, only the IR data \{$X^{I}$\} is passed to the model. CSK-Net takes a unified approach by addressing both problems concurrently rather than treating them separately. 

\subsection{CSK-Net: Spectral-based Knowledge Distillation Network} 
Semantic segmentation networks commonly employ an encoder-decoder architecture and can be denoted as $f_\theta$, where $\theta$ represents the learnable parameters. Since the shallow layers in CNNs retain the style-related information by capturing local structures \cite{zeiler2014visualizing,zheng2021deep}, CSK-Net contains shared encoders, i.e, shared convolutional layers for optical-infrared (EO-IR) images with individual batch norm layers. These shared encoder blocks consist of five stages, \{$f_{i}| n = 1,2,3,4,5$\}, where  the $i^{th}$ encoding stage is denoted as $f_i(.)$, and maps the features $F_{i-1}$ to $F_{i}$, and d(.) denotes the decoder layers, as shown in Fig. \ref{fig:MRFP}. As the model progresses through its layers, the features tend to become more specific to particular modalities. Hence, these later layers are preserved individually to uphold and retain the modality-specific semantic information.
The outputs of the decoder for both the EO and IR branches are represented as $\textbf{F}_{I}$ and $\textbf{F}_{O}$, and are fed to the Gated Spectral Unit (GSU) block as shown in Fig. \ref{fig:MRFP} (a). \\
\textbf{Mixed Feature Exchange}
This strategy serves as a self-adaptive modality fusion method, allowing for the retention of modality-specific statistics within each branch using channel and spatial exchange. Batch normalization layers are beneficial in enhancing the overall generalization capability and in preserving the distinctive characteristics of various modalities present in the data \cite{chang2019domain,wang2020learning}. 
It performs normalization on feature maps, followed by affine transformation using $\gamma$ and $\beta$ parameters. Channel exchange shows the correlation between the feature map $f_{i}$ and its corresponding scaling factor $\gamma_{i}$, showing the redundancy of feature maps on final predictions when $\gamma$ $\rightarrow$ 0, and exchange the channels of small $\gamma_{i}$  with the other sub-networks. Channel exchange \textit{C} is executed across all stages of the encoders within CSK-Net. Spatial exchange is used for exchanging the features in the dual branches specifically in the spatial dimension.
An exchange mask denoted as M $\in$ $R^{n,c,h,w}$ is generated, wherein values of 0 and 1 correspond to elements designated for non-exchange and exchange, as given below, 
\begin{equation}
M(n,c,h,w) = \left\{
        \begin{array}{ll}
           0 & \text{if} \quad \quad  w \% 2 = 0 \\
           1 & \text{if} \quad \quad  otherwise
        \end{array}
    \right.
\end{equation}
where n, c, h, and w represent the batch size, number of channels, and height width respectively. Features along the width dimension are exchanged in the dual branch only for the last two stages of the encoders.
Sparsity constraints on the $\gamma$ parameter of the batch norm have been used to automatically identify and prune during training to yield compact models\cite{liu2017learning}. During knowledge distillation, there is a continuous alteration of feature-level statistics within both the shallower and deeper sections of the models. This leads to constant fluctuations in the learnable parameters of the batch normalization, particularly the $\gamma$ parameter. Consequently, the sparsity constraints imposed during knowledge distillation are counterproductive to the learning process, enforcing stringent limitations and resulting in a decrease in performance.\\
\textbf{Gated Spectral Unit} It is proposed to enforce spectral learning, inspired by  Gated Multimodal Units \cite{arevalo2017gated}. The main idea for the multiplicative gates is to determine which input has a greater impact on generating the correct output for a rich multimodal representation. This approach avoids manual adjustments and enables the model to learn from the training data independently. It helps to learn the spectral properties from the EO and IR branches and learns to decide the influence of different unit's activation using gates.  Fig. \ref{fig:MRFP} (d) depicts the structure of a GSU. The output of the EO and IR branches and their summation are fed to the GSU block. These outputs are passed through convolution layers and then with the tanh activation function, as given below,
$$
h_{1} = tanh (W_{1} \ast F_{I} )  \eqno{(2)}  
$$
$$
h_{2} = tanh (W_{2} \ast F_{O} )  \eqno{(3)}
$$
$$
h_{3} = tanh (W_{3} \ast (F_{I} + F_{O}) )  \eqno{(4)}
$$
where, \textit{W} represents the convolution weights.
For each branch, gate neuron Z is computed, given by,
$$
Z_{1} = \sigma (W_{1} \times [F_{I},F_{O},F_{I} + F_{O}]) \eqno{(5)}
$$
where [·, ·] denotes the concatenation operator and $\sigma$ denotes sigmoid operation. The final output predictions of the fusion block $F^{fuse}$ is given by,

$$
F^{fuse} = Z_{1} \times h_{1} + Z_{2} \times h_{2} + Z_{3} \times h_{3}  \eqno{(6)}
$$
where ($\times$) represents the multiplication operation. 
CSK-Net model has three outputs as shown in the figure, two for each modality, i.e., for optical and IR, and the third output from the GSU block, given by $F^{fuse}$. GSU is applied during both the training phase of the model and during inference for multi-modal scenarios. However, in the case of missing modality scenarios, it is omitted.
GSU helps in preserving the high-frequency, modality-specific information within the IR branch of the model during distillation, by preventing forced feature alignment between different modalities.

\subsection{Training scheme}


The training process consists primarily of two key steps: (1) pre-training the baseline DeepLabV3+ segmentation model on optical images, and (2) training the CSK-Net model using EO-IR coregistered images while concurrently distilling optical knowledge from the pre-trained model into the optical branch of CSK-Net.
During the first training step, the baseline segmentation model is trained with the optical images with their corresponding labels \{$X^{O}, Y$\}, using the 
$L_{ST1}(p,y)$ segmentation loss to train the whole model. The segmentation loss $L_{seg}(p,y)$ consists of the summation of cross-entropy and dice loss, and is given by,
$$
L_{seg}(p,y)= - \sum_{i} y_{i}log(p_{i}) + 1 - \frac{2\sum_{i}p_{i}y_{i}}{\sum_{i}y_{i}+\sum_{i}p_{i}}\ \eqno{(7)}
$$
where, \textit{y} and \textit{p} denote the ground truth labels and the pixel-wise predictions respectively.

During the second training step, the CSK-Net model is trained, and distillation from the pre-trained optical model is performed using two distillation loss terms, namely $L_{D1}$ and $L_{D2}$. Using the multi-class pixel-wise predictions obtained by the pre-trained model denoted as $p^{PO}$, the distillation loss $L_{D1}$ is given by,
$$
L_{D1} (p, p^{PO}) = \sum_{i}p_{i}^{PO} log\frac{p_{i}^{PO}}{p_{i}} -\sum_{i}p_{i}^{PO} log (p_{i})  \ \eqno{(8)}
$$
where \textit{p} represents the predictions made by the optical branch of CSK-Net. 
Kullback-Leiber (KL) divergence along with the cross-entropy loss is used to generate similar predictions made by the pre-trained optical branch. 

Inspired by the deep distillation strategy from \cite{xie2015holistically}, the multi-level semantic information is distilled from the last two layers of the encoder  of the pre-trained model to the last two layers of the shared encoders of the optical branch of CSK-Net, using the mean square error loss, given by,
$$
L_{D2} (F, F^{PO}) = \sum_{i \in \{4,5\}}\|F_{i} - F^{PO}_{i}\|_2 + \|F_{d} - F^{PO}_{d}\|_2 \ \eqno{(9)}
$$
which measures the difference between the features of the last two layers of the encoders and the decoder output. 

To train the shared encoders for superior modality-specific style representation, contrastive learning $L_{CL}$ \cite{gutmann2010noise} is used for the last four layers of the encoders to improve the intra-domain mining. 
The features are mapped into an embedding space using a projection head $h_{pixel}$ to promote discriminative feature learning.
This process aims to bring pixel embeddings from the same category closer together while pushing pixel embeddings from different categories farther apart. Using the pixel-wise labels, pixels belonging to the same class are treated as positive samples, while those belonging to different classes are considered negative samples. The pixel-wise contrastive loss is formulated as,
$$
L_{CL} = -\sum_{C(i)=C(j)}log\frac{r(e_{i},e_{j})}{\sum_{k=1}^{Np}r(e_{i},e_{j})}\ \eqno{(10)}
$$
where, \textit{$e_{i}$} represents the $i^{th}$ feature map obtained from the projection head, \textit{Np} stands for the total number of pixels, \textit{r} denotes the similarity measure. Similarity is calculated using the exponential similarity function: r($e_{i},e_{j}$) = exp(\textit{s}($e_{i},e_{j}$) / $\tau$), where \textit{s} represents the cosine similarity, and $\tau$ is the temperature parameter. A semi-hard example sampling strategy is adopted, where the negative samples are retained from the whole training batch, with the top 10\% nearest negatives and farthest positives selected for each anchor sampling \cite{chen2023pipa}. The joint loss function for training CSK-Net is given by, 
$$
L_{ST2} = L_{seg}(y,p_{(F,IR)})
 + L_{D1} (p,p^{PO}) + L_{D2}(F,F^{PO}) + L_{CL} \ \eqno{(11)}
$$
where y denotes the ground truth labels,  $L_{seg}$ is used to optimize the model's output from fused segmentation predictions $p_{F}$ generated using GSU and the IR segmentation head predictions $p_{IR}$, along with contrastive loss $L_{CL}$. During inference, for multi-modal settings, the final predictions from the GSU block are used. However, in scenarios where a modality is missing, only the IR branch of CSK-Net is employed during inference. This IR branch has an identical configuration to that of DeepLabV3+.

\begin{figure*}
    \centering
    \includegraphics[scale=0.085]{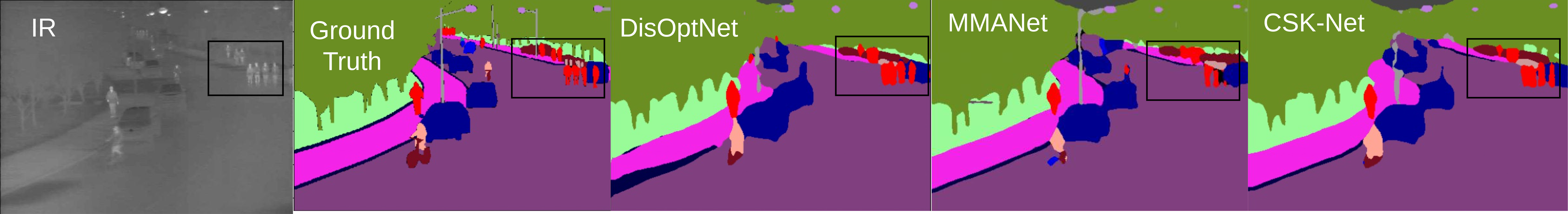}
    \caption{Comparison of output predictions of CSK-Net with baseline and state-of-the-art models on MVSS dataset for missing modality scenario. The cluttered car labels (blue) can be seen in, as opposed to the ground truth labels.  CSK-Net is able to segment cars more accurately. Our model is also able to predict bicycles (light brown) more accurately.}
    \label{fig:comparison1}
\end{figure*}

\begin{figure*}
    \centering
    \includegraphics[scale=0.21]{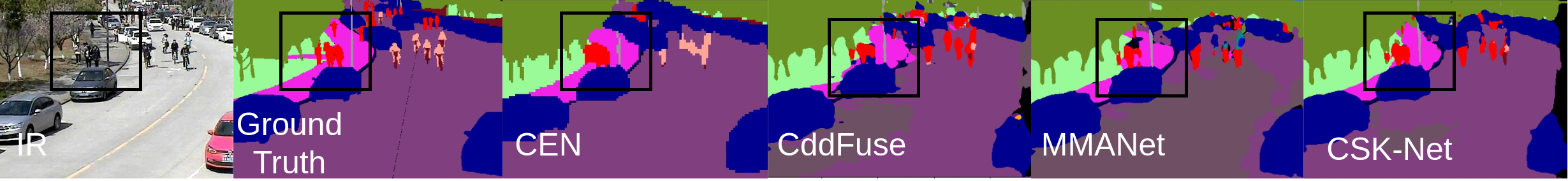}
    \caption{Comparison of output predictions of CSK-Net with baseline and state-of-the-art models on MVSS dataset for multi-modality scenario.}
    \label{fig:comparison2}
\end{figure*}

\section{Experimental Results}

\subsection{Setup}
This section contains the experimental details and configurations of CSK-Net and other state-of-the-art models for the downstream task of semantic segmentation for both multi-modal settings with optical (EO) and IR data and for missing modality scenarios where only IR data is available. 

\textbf{Datasets and metrics}
Three public datasets are used for benchmarking, i.e., MSRS, MVSS, and FMB datasets. MSRS dataset contains 1,444 high-quality image pairs with a resolution of 480 × 640 containing 9 classes such as color cones, cars, bikes, and pedestrians. The dataset is randomly split, with 1083 samples for training and 361 for testing.  The MVSS dataset contains images of urban scenes, with both daytime and nighttime conditions with 26 classes such as cars, buses, poles, buildings, and pedestrians. It contains a total of 1616 samples, with 1004 samples used for training and 612 samples for testing. The images are re-sized to 320 $\times$ 480. FMB dataset contains 1500 infrared and visible image pairs with 14 categories like Road, Sidewalk, Building, and Motorcycle for a wide variety of environmental conditions like fog, heavy rain, and low-light conditions.    To quantitatively measure the segmentation performance of the models, Intersection over Union is used as a metric.

\begin{table}[]
\begin{tabular}{|c|l|l|l|}
\hline
\textbf{\begin{tabular}[c]{@{}c@{}}Multi-Spectral\\ Settings\end{tabular}} & \textbf{Method}                                                                    & \textbf{mIoU}                                                         & \textbf{\#Params}                                                             \\ \hline
\textbf{Baseline}                                                          & \begin{tabular}[c]{@{}l@{}}VI\\ IR\end{tabular}                                    & \begin{tabular}[c]{@{}l@{}}48.89\\ 42.82\end{tabular}                 & \begin{tabular}[c]{@{}l@{}}11.68 M\\ 11.68 M\end{tabular}                     \\ \hline
\textbf{\begin{tabular}[c]{@{}c@{}}Multi-modal\\ Methods\end{tabular}}     & \begin{tabular}[c]{@{}l@{}}C.E.N.\\ CDDFuse\\ MMANet\\ CSK-Net (Ours)\end{tabular} & \begin{tabular}[c]{@{}l@{}}51.33\\ 48.41\\ \underline{49.31}\\ \textbf{51.65}\end{tabular} & \begin{tabular}[c]{@{}l@{}}99.13 M\\ 13.47 M\\ 71.70 M\\ 14.24 M\end{tabular} \\ \hline
\textbf{\begin{tabular}[c]{@{}c@{}}Missing\\ Modality\end{tabular}}        & \begin{tabular}[c]{@{}l@{}}DisOptNet\\ MMANet\\ CSK-Net (Ours)\end{tabular}        & \begin{tabular}[c]{@{}l@{}}43.22\\ \textbf{46.83}\\ \underline{46.40}\end{tabular}         & \begin{tabular}[c]{@{}l@{}}11.68 M\\ 88.01 M\\ 11.68 M\end{tabular}           \\ \hline
\end{tabular}
\caption{Performance comparison of IoU (\%) of proposed CSK-Net with other state-of-the-art models on MVSS dataset.}
\label{MVSS results}
\end{table}

\begin{table}[]
\begin{tabular}{|c|l|c|l|}
\hline
\textbf{\begin{tabular}[c]{@{}c@{}}Multi-Spectral\\ Settings\end{tabular}} & \textbf{Method}                                                                    & \multicolumn{1}{l|}{\textbf{Publication}}                                      & \textbf{mIoU}                                                         \\ \hline
\textbf{Baseline}                                                          & \begin{tabular}[c]{@{}l@{}}VI\\ IR\end{tabular}                                    & \begin{tabular}[c]{@{}c@{}}-\\ -\end{tabular}                                  & \begin{tabular}[c]{@{}l@{}}49.72\\ 45.95\end{tabular}                 \\ \hline
\textbf{\begin{tabular}[c]{@{}c@{}}Multi-modal\\ Methods\end{tabular}}     & \begin{tabular}[c]{@{}l@{}}C.E.N.\\ CDDFuse\\ MMANet\\ CSK-Net (Ours)\end{tabular} & \begin{tabular}[c]{@{}c@{}}TPAMI 2022\\ CVPR 2023\\ CVPR 2023\\ -\end{tabular} & \begin{tabular}[c]{@{}l@{}}49.01\\ 48.41\\ \underline{53.97}\\ \textbf{54.37}\end{tabular} \\ \hline
\textbf{\begin{tabular}[c]{@{}c@{}}Missing\\ Modality\end{tabular}}        & \begin{tabular}[c]{@{}l@{}}DisOptNet\\ MMANet\\ CSK-Net (Ours)\end{tabular}        & \begin{tabular}[c]{@{}c@{}}TGRSS 2022\\ CVPR 2023\\ -\end{tabular}             & \begin{tabular}[c]{@{}l@{}}46.34\\ \textbf{48.59}\\ \underline{48.20}\end{tabular}         \\ \hline
\end{tabular}
\caption{Performance comparison of Intersection over Union (IoU\%) of proposed CSK-Net with other state-of-the-art models on FMB  dataset. CSK-Net outperforms multi-modal, especially in challenging conditions of low light. It performs comparable to MMANet, with significantly fewer parameters.}
\label{FMB results}
\end{table}




\subsubsection{Implementation details}

All experiments are conducted using three NVIDIA Quadro RTX 5000 GPUs, and all models are trained with a batch size of 8 for 200 epochs. DeepLabV3+ \cite{chen2018encoder} is the baseline segmentation model with the EfficientNet-B3 \cite{tan2019efficientnet}  backbone, which is pre-trained on ImageNet. All models have horizontal flips as the data augmentation with 50\% probability. The models are trained using an SDG optimizer, with an initial rate of 5 $\times$ $10^{-3}$. A polynomial scheduler decreases the learning rate after each epoch, a decay factor of (1 - step/total steps)$^{0.9}$. All state-of-the-art models are implemented using their official open-source codes.

\subsection{Comparison with SOTA methods}
The performance of CSK-Net for multi-modal fusion is compared with other state-of-the-art fusion techniques like Channel Exchanging Network (C.E.N) \cite{wang2022channel} and CDDFuse. The performance for missing modality scenarios, where the models are trained using both optical and Infrared (IR) modalities, however for inference, only the IR sensor data is available for inference, is compared with other state-of-the-art models like DisOptNet and MMANet.  To ensure fair comparisons, the baseline models DeepLabV3+, and DisOptNet are re-implemented using the same training strategy as CSK-Net. Additionally, C.E.N is re-implemented according to the training strategy as specified in \cite{wang2022channel} with a batch size of 8, for 200 epochs.

\subsubsection{Quantitative evaluation}

Tables \ref{MVSS results}, \ref{FMB results}, and \ref{MSRS results} show the segmentation performance of CSK-Net and other state-of-the-art models. The performance of the baseline DeepLabV3+ model for Oracle settings is shown, where it is trained and tested solely on visible (VI) or infrared (IR) data.
To evaluate the performance when one modality is missing, all other models are trained using pairs of electro-optical (EO) and infrared (IR) data but are exclusively tested on IR data.
Segmentation performance for multi-modal fusion is shown in \textit{Multi-modal Methods}, where CSK-Net consistently outperforms image-based fusion model CDDFuse by 3.24 \% on the MVSS dataset, 5.96\% on the FMB dataset, and 3.81 \% on the MSRS dataset. It also outperforms the C.E.N model by 2.84 \% on average across all datasets, with only 12 \% of its total number of parameters. For the \textit{missing modality scenarios}, as shown in the tables, CSK-net outperforms DisOptNet by 3.18 \% on MVSS, 1.85 \% on the FMB dataset, and 2\% on the MSRS dataset.  It outperforms MMANet for both tasks on all the datasets by an average of 4.775\%, with significantly fewer parameters.

\begin{table}[]
\begin{tabular}{|c|l|l|l|}
\hline
\textbf{\begin{tabular}[c]{@{}c@{}}Multi-Spectral\\ Settings\end{tabular}} & \textbf{Method}                                                                    & \textbf{mIoU}                                                         & \textbf{\#Params}                                                             \\ \hline
\textbf{Baseline}                                                          & \begin{tabular}[c]{@{}l@{}}VI\\ IR\end{tabular}                                    & \begin{tabular}[c]{@{}l@{}}64.33\\ 61.69\end{tabular}                 & \begin{tabular}[c]{@{}l@{}}11.68 M\\ 11.68 M\end{tabular}                     \\ \hline
\textbf{\begin{tabular}[c]{@{}c@{}}Multi-modal\\ Methods\end{tabular}}     & \begin{tabular}[c]{@{}l@{}}C.E.N.\\ CDDFuse\\ MMANet\\ CSK-Net (Ours)\end{tabular} & \begin{tabular}[c]{@{}l@{}}61.04\\ \underline{66.71}\\ 66.24\\ \textbf{69.38}\end{tabular} & \begin{tabular}[c]{@{}l@{}}99.13 M\\ 13.47 M\\ 71.70 M\\ 14.24 M\end{tabular} \\ \hline
\textbf{\begin{tabular}[c]{@{}c@{}}Missing\\ Modality\end{tabular}}        & \begin{tabular}[c]{@{}l@{}}DisOptNet\\ MMANet\\ CSK-Net (Ours)\end{tabular}        & \begin{tabular}[c]{@{}l@{}}\underline{63.84}\\ 61.34\\ \textbf{65.83}\end{tabular}         & \begin{tabular}[c]{@{}l@{}}11.68 M\\ 88.01 M\\ 11.68 M\end{tabular}           \\ \hline
\end{tabular}
\caption{Performance comparison of Intersection over Union
(IoU\%) of proposed CSK-Net with other state-of-the-art models
on  MSRS dataset.}
\label{MSRS results}
\end{table}

\begin{table}[]
\centering
\begin{tabular}{|c|c|}
\hline
\textbf{Method}                                                                                                 & \textbf{mIoU}                       \\ \hline
CSK-Net                                                                                                & \textbf{69.38}                      \\ \hline
CSK-Net w/o Contrastive Learning                                                                       & 68.01                      \\ \hline
\multicolumn{1}{|l|}{CSK-Net w/o Gated Spectral Fusion}                                                 & \multicolumn{1}{l|}{68.90} \\ \hline
CSK-Net w/o Mixed Feature Exchange                                                                     & 68.54                      \\ \hline
\begin{tabular}[c]{@{}c@{}}CSK-Net w/o Contrastive Learning w/o \\ Mixed Feature Exchange\end{tabular} & 68.37                      \\ \hline
\end{tabular}
\caption{Ablation for contrastive learning and mixed feature exchange on CSK-Net for multi-modal setting on MSRS dataset.}
\label{table:ablation1}
\end{table}

\begin{figure*}
    \centering
    \includegraphics[scale=0.30]{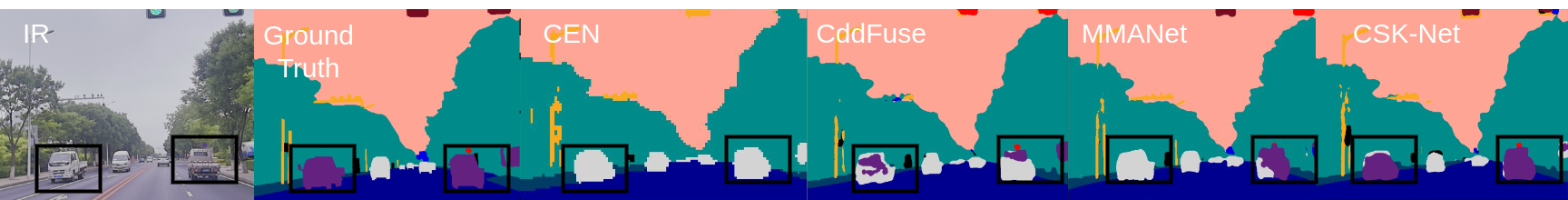}
    \caption{Comparison of output predictions of CSK-Net with baseline and state-of-the-art models on FMB dataset for multi-modal scenario.}
    \label{fig:comparison3}
\end{figure*}

\begin{figure*}
    \centering
    \includegraphics[scale=0.30]{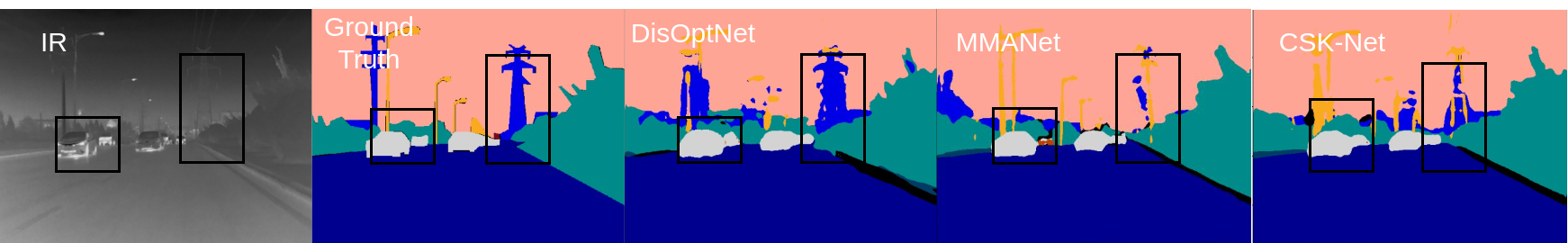}
    \caption{Comparison of output predictions of CSK-Net with baseline and state-of-the-art models on FMB dataset for missing modality scenario.}
    \label{fig:comparison4}
\end{figure*}




\subsubsection{Qualitative evaluation}
The qualitative comparisons are shown in Fig. \ref{fig:comparison1}, \ref{fig:comparison2}, \ref{fig:comparison3} and \ref{fig:comparison4}. 
Compared to contemporary models, the predictions made by CSK-Net for multi-modal fusion settings are superior, especially to image-based fusion models for low-light and foggy conditions. For the missing modality scenario,  CSK-Net performs feature distillation from optical images and acquires domain-invariant features across various spectra for the same object categories. This contributes to the model's ability to make superior predictions.

\subsection{Ablation Study}
Ablation experiments are conducted to verify the significance of different components of CSK-Net. All experiments conducted have the same training settings mentioned in the implementation details. Table \ref{table:ablation1} shows the effectiveness of the mixed feature exchange strategy, GSU, and contrastive learning. The table shows that the mixed feature exchange strategy aids the distillation, to avoid forced feature alignment.  Contrastive learning assists in preserving the style information during the distillation process. Combining all the techniques yield the most optimal overall performance.

\section{Conclusions}
This paper introduces a novel multi-modal fusion approach known as CSK-Net for multi-spectral semantic segmentation tasks, using spectral-based knowledge distillation, for both multi-modal and missing modality scenarios. 
The model contains shared encoders with individual batch norms for both modalities trained without any sparsity constraints. Spectral knowledge distillation is used to distill optical knowledge from the pre-trained baseline segmentation model to the optical branch of CSK-Net. 
Pixel-wise contrastive loss is used to train the encoders, to retain modality-shared information, and Gated Spectral Fusion (GSU) and mixed feature exchange are proposed to regulate the correlation between low and high-frequency information during the knowledge distillation. CSK-Net consistently achieves superior performance on three public benchmarking datasets. 
It outperforms CDDFuse for multi-modal fusion by \textbf{4.33\%} on average across all three datasets and outperforms MMANet by \textbf{5.88\%} for multi-modal setting and \textbf{3.67\%} for missing modality setting on average across all three datasets.